\documentclass[journal]{IEEEtran}

\usepackage{xcolor,soul,framed} 

\colorlet{shadecolor}{yellow}
\usepackage[pdftex]{graphicx}
\graphicspath{{../pdf/}{../jpeg/}}
\DeclareGraphicsExtensions{.pdf,.jpeg,.png}

\usepackage[cmex10]{amsmath}
\usepackage{array}
\usepackage{mdwmath}
\usepackage{mdwtab}
\usepackage{eqparbox}
\usepackage{url}
\usepackage{amsthm}
\newtheorem{definition}{Definition}
\usepackage{amssymb}
\usepackage{amsmath, amssymb}
\usepackage{algorithm}
\usepackage{algpseudocode}
\usepackage{amsmath}
\usepackage{float}
\usepackage{multirow}
\usepackage{cite}
\usepackage{xcolor}  
\usepackage[
    colorlinks=true,        
    linkcolor=red,          
    citecolor=blue,         
    urlcolor=blue,         
    filecolor=magenta       
]{hyperref}

\begin{document}
\bstctlcite{IEEEexample:BSTcontrol}
    \title{DoGCLR: Dominance-Game Contrastive Learning Network for Skeleton-Based Action Recognition}
  \author{Yanshan Li,
      Ke Ma,
      Miaomiao Wei,
      Linhui Dai$^{*}$

  \thanks{This research was partially supported by the National Natural Science Foundation of China (Grant No. 62076165), the Innovation Team Project of the Department of Education of Guangdong Province (Grant No. 2020KCXTD004), the Educational Science Planning Project of Guangdong Province (Grant No. 2022GXJK367), and the Scientific Foundation for Youth Scholars of Shenzhen University, China.}

  \thanks{Yanshan Li, Ke Ma, Miaomiao Wei, and Linhui Dai$^{*}$ are with Shenzhen University, the Institute of Intelligence Information Processing, and the Guangdong Key Laboratory of Intelligent Information Processing, Shenzhen 518000, China (emails: lys@szu.edu.cn; 2400042033@email.szu.edu.cn; 1054772440@qq.com; dailinhui@szu.edu.cn).}
  }

\markboth{IEEE TRANSACTIONS ON PATTERN ANALYSIS AND MACHINE INTELLIGENCE, VOL.~xxx, NO.~xxx, xxx
}{Roberg \MakeLowercase{\textit{et al.}}: High-Efficiency Diode and Transistor Rectifiers}

\maketitle

\begin{abstract}
Existing self-supervised contrastive learning methods for skeleton-based action recognition often process all skeleton regions uniformly, and adopt a first-in-first-out (FIFO) queue to store negative samples, which leads to motion information loss and non-optimal negative sample selection. To address these challenges, this paper proposes Dominance-Game Contrastive Learning network for skeleton-based action Recognition (DoGCLR), a self-supervised framework based on game theory. DoGCLR models the construction of positive and negative samples as a dynamic Dominance Game, where both sample types interact to reach an equilibrium that balances semantic preservation and discriminative strength. Specifically, a spatio-temporal dual-weight localization mechanism identifies key motion regions and guides region-wise augmentations to enhance motion diversity while maintaining semantics. In parallel, an entropy-driven dominance strategy manages the memory bank by retaining high-entropy (hard) negatives and replacing low-entropy (weak) ones, ensuring consistent exposure to informative contrastive signals. Extensive experiments are conducted on NTU RGB+D and PKU-MMD datasets. On NTU RGB+D 60 X-Sub/X-View, DoGCLR achieves 81.1\%/89.4\% accuracy, and on NTU RGB+D 120 X-Sub/X-Set, DoGCLR achieves 71.2\%/75.5\% accuracy, surpassing state-of-the-art methods by 0.1\%, 2.7\%, 1.1\%, and 2.3\%, respectively. On PKU-MMD Part I/Part II, DoGCLR performs comparably to the state-of-the-art methods and achieves a 1.9\% higher accuracy on Part II, highlighting its strong robustness on more challenging scenarios.
\end{abstract}


%
\IEEEpeerreviewmaketitle


\section{Introduction}

\IEEEPARstart{I}{n} recent years, skeleton-based action recognition has attracted extensive attention from both academia and industry \cite{Ji2021ArbitraryView, Cheng2022CrossModality}. Owing to its compact representation and strong robustness against appearance variations, skeleton data provide a high-level abstraction of human motion. However, supervised learning approaches for this task often rely on large-scale labeled datasets, which are costly and time-consuming to construct. To overcome this limitation, self-supervised learning (SSL) has emerged as a promising alternative \cite{Su2019PredictCluster, Lin2020MS2L, Zheng2018Unsupervised, Thoker2021SkeletonContrastive} enabling representation learning without manual annotations. Among various approaches, contrastive learning has demonstrated remarkable effectiveness for skeleton-based action recognition due to its intuitive learning principle and strong discriminative capability. Contrastive learning learns by contrasting positive and negative sample pairs, maximizing agreement between different augmented views of the same sample while minimizing similarity between different samples. Through this process, the model acquires invariant and discriminative representations that generalize well to downstream tasks \cite{He2019MoCo}.

Despite recent progress, existing self-supervised contrastive learning methods for skeleton-based action recognition still suffer from non-optimal construction of positive samples and management of negative samples. Specifically, the quality and value of positive and negative samples have not been precisely optimized, limiting the effectiveness of feature learning and the model’s ability to refine discriminative boundaries. The main challenges are as follows:

(1) \textbf{Positive sample construction.} Traditional methods \cite{He2019MoCo, Guo2022AimCLR, Lin2023ActionletDependent, Gao2021ContrastiveSSL} typically apply global and uniform data augmentation and feature extraction strategies to skeleton sequences. These approaches fail to leverage the structural property that “the degree of skeletal joints reflects their contribution to the overall action,” nor do they distinguish between key motion regions and non-critical areas. Consequently, the resulting positive samples often suffer from semantic distortion or information dilution, making it difficult to simultaneously ensure high semantic integrity and high information diversity, thereby constraining the model’s ability to effectively capture core motion features.

(2) \textbf{Negative sample management.} Previous approaches \cite{Guo2022AimCLR, Lin2023ActionletDependent, Gao2021ContrastiveSSL, Guo2024AimCLRPlus} generally adopt a first-in-first-out (FIFO) mechanism in their negative sample memory banks without quantifying the information value of each negative sample. As a result, the memory bank often includes numerous low-value samples. Some negatives differ greatly from positives and contribute little to boundary refinement, whereas others closely resemble positives and may mislead representation learning. These low-value samples occupy valuable queue capacity and prevent the retention of high-value hard negatives that are crucial for discriminative learning. This lack of strong contrastive constraints ultimately hampers the model’s ability to accurately delineate feature boundaries, leading to high misclassification rates among similar actions.

To address these issues, we argue that sample construction and selection in contrastive learning can be viewed as a dynamic optimization process — one that naturally aligns with principles from game theory. In such a setting, positive and negative samples act as competing agents: positives seek to strengthen semantic alignment, while negatives aim to sharpen discriminative boundaries. The Dominance Game (DG) provides an ideal theoretical framework for modeling this competition, ensuring equilibrium stability under controlled perturbations.

Building on this insight, we propose Dominance-Game Contrastive Learning network for skeleton-based action Recognition (DoGCLR), a self-supervised framework that integrates game-theoretic optimization into both positive and negative sample learning. DoGCLR establishes a dual-dimensional dominance system that jointly enhances positive sample representativeness and negative sample selectivity. 

The major contributions are summarized as follows:

(1) \textbf{A novel Dominance-Game-based contrastive learning network.} The proposed DoGCLR establishes a dual-dimensional dominance system encompassing positive sample optimization and negative sample selection. Through accurate key-region localization, region-wise differentiated augmentation, and entropy-prioritized negative sampling, the network strengthens both the representativeness of positive samples and the constraint strength of negative samples. Combined with a contrastive loss optimization module, it significantly enhances recognition accuracy, generalization, and training efficiency.

(2) \textbf{Dominant positive sample construction via spatio-temporal dual-weight localization.} The Spatio-temporal Dual-Weight synergistic localization algorithm-based Key Region partitioning Module (DW-KRM) partitions key motion regions by combining spatial and temporal importance obtained from the Joint-Degree Activation Module (JDAM). Based on this localization, the Dual-scale-Game-based partitioned data Augmentation module (DGA) applies strong augmentations to key regions and weak ones to non-key regions, maintaining semantic consistency while enhancing motion diversity. This cooperative optimization allows the constructed positive samples to be both informative and highly representative.

(3) \textbf{Entropy-driven Dominance Game for negative sample management.} An entropy-prioritized memory bank is designed, where the dominance strategy in game theory is coupled with negative sample valuation. The Entropy-driven Dominance Game Replacement Queue (EDGRQ) quantifies each sample’s dominance via information entropy, retaining high-entropy (hard) negatives and discarding low-entropy (weak) ones. This transforms negative sample management from “uniform processing” to “dynamic dominance-based selection,” ensuring the model consistently learns from the most informative contrastive signals.

(4) \textbf{Comprehensive experiment results.} Extensive experiments demonstrate the superior performance of DoGCLR across multiple benchmarks. On NTU RGB+D 60/120, DoGCLR consistently outperforms baselines under both X-Sub and X-View/X-Set protocols, improving accuracy by 2–6\% and surpassing SOTA by up to 3\%. On PKU-MMD, it achieves comparable or superior results, with a 1.9\% gain on Part II, validating its robustness in complex multi-person scenarios. Our code is available at \url{https://github.com/Ixiaohuihuihui/DoGCLR}.

\section{Related Work}

\subsection{Skeleton Key-Region Modeling and Spatio-Temporal Augmentation Methods}

In recent years, self-supervised skeleton-based action recognition has achieved remarkable progress. The core idea is to learn more discriminative latent representations by constructing diverse spatio-temporal transformations or auxiliary tasks. In 2020, Lin et al. \cite{Lin2020MS2L} proposed MS2L, a multi-task self-supervised learning framework that jointly performs “motion prediction + jigsaw puzzle” recognition to capture temporal dependencies and motion semantics, thereby significantly enhancing the representation capacity of unlabeled skeleton data. Subsequently, Li et al. \cite{Li2020ShapeMotion} introduced STVIM, which realizes view-invariant shape and motion representations under a geometric-algebra framework through rotor-based viewpoint transformations. This approach provides a structural-invariance prior that supports the stability of key-region enhancement. In 2021, Rao et al. \cite{Rao2021MomentumLSTM} integrated multiple data-augmentation strategies into AS-CAL to maintain feature stability under viewpoint variations and temporal perturbations. Meanwhile, Li et al. \cite{Li2021CrossViewConsistency} proposed CrosSCLR, which employs complementary supervision across multiple views to enforce Cross-View consistency learning. Guo et al. \cite{Guo2022AimCLR} further developed AimCLR, which enhances feature discriminability through extreme augmentations, Energy-Attention-Guided Dropping (EADM), Dual-Distribution Divergence Minimization (D3M), and Nearest-Neighbor Mining (NNM). In the same year, Liu et al. \cite{Liu2021AdaptiveMultiView} presented AMW-GCN, which models both spatial structures and temporal dynamics via an adaptive multi-view transformation module combined with a multi-stream graph-convolutional structure, thereby improving the robustness of action representation under different observation viewpoints. In 2022, Chen et al. \cite{Chen2022MixedSkeleton} incorporated topological information to design a hybrid-augmentation contrastive-learning network that better mines hard samples. Xia et al. \cite{Xia2022LAGANet} proposed LAGA-Net, which employs motion-guided channel attention and spatio-temporal attention to simultaneously strengthen frame-level saliency and global dependency — an idea consistent with region-saliency-driven augmentation. In 2023, Dong et al. \cite{Dong2023HiCLR} introduced HiCLR, which performs multi-scale feature fusion through progressive down-sampling in the encoder stage, while Lin et al. \cite{Lin2023ActionletDependent} proposed ActCLR, employing node masking and frame expansion augmentation to “break” structural consistency and thereby explore richer motion patterns at finer granularity. In 2024, Wu et al. \cite{Wu2024SCDNet} developed SCD-Net, which integrates spatial- and temporal-clue disentanglement modules to further enhance the semantic expressiveness of action representations. Unlike previous methods, DoGCLR establishes a spatio-temporal dual-dimensional local DG framework to locate key motion regions and models a global DG to achieve targeted skeleton augmentation.

\subsection {Negative Sample Management and Dynamic Memory Mechanism Optimization}

The selection and maintenance of negative samples are crucial factors influencing the performance of contrastive learning. In 2019, He et al. \cite{He2019MoCo} proposed MoCo, which employs a momentum encoder and a FIFO queue to construct a large-scale negative sample memory, laying the foundation for self-supervised negative sample learning. In 2021, Li et al. \cite{Li2021CrossViewConsistency} introduced multi-view complementary supervision in CrosSCLR, which improves the discriminability of negative samples by leveraging view-level consistency. In 2023, Hua et al. \cite{Hua2023PartAwareCLR} proposed Part-Aware Contrastive Learning, where skeleton partitioning is utilized for hard-sample sampling to enhance the discriminability of local features. Wang et al. \cite{Wang2023SS3D} presented ASAR, which employs an adaptive re-calibration strategy to balance the distribution of negative samples, thereby mitigating representation collapse and gradient imbalance. In 2024, Guo et al. \cite{Guo2024AimCLRPlus} proposed 3s-AimCLR++, which integrates tri-stream aggregation and multi-stream interaction mechanisms to more effectively exploit cross-modal and cross-channel complementary information. Different from existing approaches, DoGCLR constructs an entropy-driven DG framework that preserves hard negative samples, ensuring the network continuously learns the most informative negative representations.

\subsection{Game-Theory-Driven Self-Supervised Contrastive Learning Framework}

Game theory has been introduced into deep learning to describe the process of deriving optimal strategies in non-cooperative competition. In 2022, Li et al. \cite{Li2022MDLBP} proposed MDLBP, which characterizes local differences through multidimensional geometric relationships, reflecting the idea of coexistence between competition and cooperation in the feature space. Mao et al. \cite{Mao2022CMD} developed CMD, where a bidirectional distillation mechanism is used to construct a collaborative game relationship between different modalities. In 2023, Dong et al. \cite{Dong2023HiCLR} introduced a hierarchical contrastive mechanism in HiCLR, allowing features at different scales to be progressively optimized through strategic competition. Yu et al. \cite{Yu2024GeoExplainer} proposed GeoExplainer, which integrates the spiral theory and geometric masking mechanism to achieve interpretable modeling for spatio-temporal graph convolutional networks. By employing a PMI constraint to generate geometric perturbation masks, this work provided a new geometric perspective for self-explainable learning under game-theoretic principles. In 2024, Li et al. \cite{Li2024BICAM} proposed BI-CAM, which models explainable heat distributions through positive and negative mutual information, essentially representing a process of local game between cooperation and competition. In the same year, they further proposed GT-CAM \cite{Li2024GTCAM}, which incorporates the Shapley value allocation mechanism from cooperative game theory into feature interpretability analysis to depict the cooperative and competitive relationships among nodes within a structure. Similar to previous methods, DoGCLR aims to further enhance network performance by integrating classical game-theoretic principles into the contrastive learning framework.


\section{Method}

\subsection{Overview of the DoGCLR Network}

\begin{figure*}
  \begin{center}
  \includegraphics[width=\textwidth]{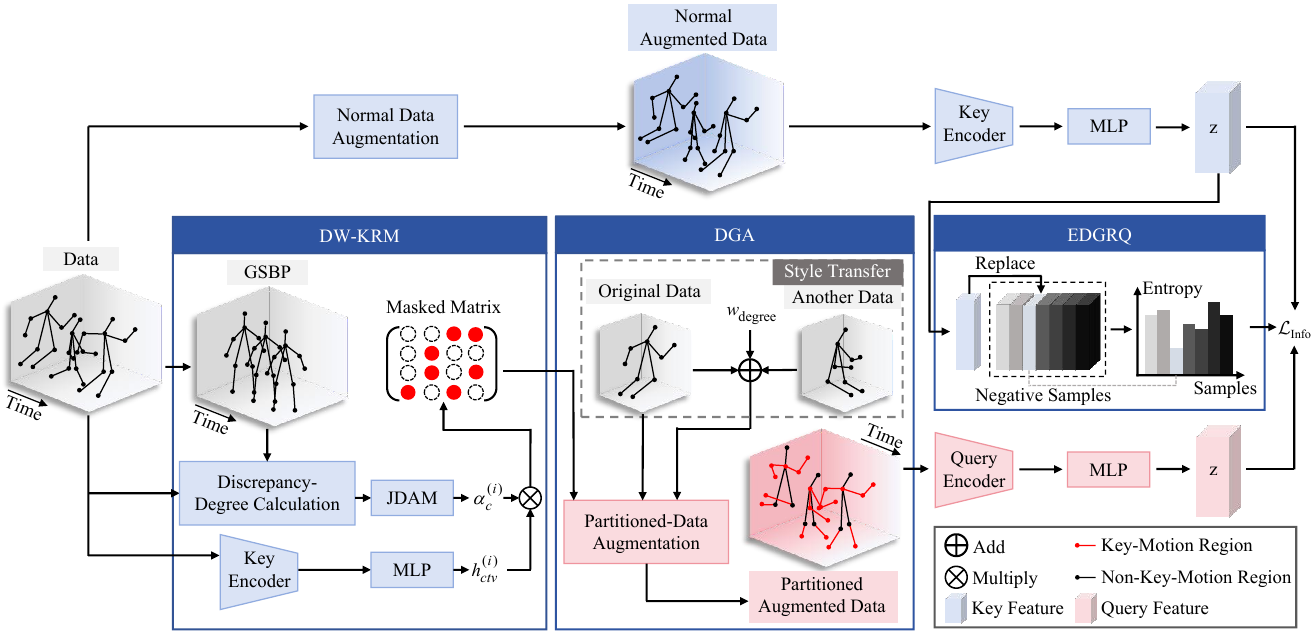}\\
  \caption{Pipeline of DoGCLR. The positive sample optimization part consists of DW-KRM and DGA, which locate the key motion regions of samples and perform partitioned data augmentation to explore richer motion patterns while maintaining semantic consistency. The negative sample selection part is composed of EDGRQ, which improves the value of negative samples by retaining hard negatives. DW-KRM, DGA and EDGRQ together constitute the dual-dimensional dominance system of “positive sample optimization and negative sample selection”.}\label{fig: 01}
  \end{center}
\end{figure*}

Dominance Game serves as the theoretical foundation of the proposed DoGCLR network. To facilitate the explanation of the network structure, the definition of the \textit{Dominance Game (DG)} is given as follows:

\begin{definition}[Dominance Game, DG]
Let $\Gamma = (N, (C_i)_{i \in N}, (u_i)_{i \in N})$ denote a game, where $N$ represents the set of players, $C_i$ represents the set of strategies of player $i$, and $u_i$ represents the payoff function of player $i$. If there exists a strategy profile $c = (c_1, \cdots, c_{|N|})$, where $c_i \in C_i$, and $|N|$ denotes the number of elements in the set $N$, such that for any player $i \in N$, any $d_i \in C_i$, $u_i(c) > u_i(d)$, where $d = (d_1, \cdots, d_{|N|})$, then $c$ is called a \textit{dominance strategy profile}, and the game is referred to as a \textit{Dominance Game (DG)}.
\end{definition}

A DG always possesses an optimal strategy under any condition, and when small perturbations occur, the strategy remains optimal, demonstrating DG’s strong robustness.

The proposed DoGCLR network follows the learning paradigm of MoCov2 \cite{He2019MoCo} and constructs a dual-dimensional dominance system consisting of positive-sample optimization and negative-sample selection. In the positive-sample construction stage, a Dual-Scale Dominance Game based on the Spatio-Temporal-Dual-weight Synergistic Localization Algorithm is proposed. This process involves two core modules: the Spatio-temporal Dual-Weight synergistic localization algorithm-based Key Region partitioning Module (DW-KRM) and the Dual-scale-Game-based partitioned data Augmentation module (DGA). Based on the optimal strategy derived from the game, targeted data augmentation is designed to explore more diverse motion patterns while maintaining semantic consistency. In the negative sample management, an Entropy-driven Dominance Game Replacement Queue (EDGRQ) is introduced. According to the optimal strategy of the game, high-entropy hard negatives are preferentially retained, while low-entropy inferior samples are discarded, providing stronger contrastive constraints for model training.

Specifically, in the DW-KRM, the Global Statistical Benchmark Pose (GSBP) is computed first. The cosine similarity between the GSBP and the input skeleton data is then calculated, and the negative gradient of the node features with respect to this similarity yields the Discrepancy-Degree (DD) of each node. Subsequently, in the Joint-Degree Activation Module (JDAM), the adjacency matrix of the skeleton graph is used to count the number of edges connected to each node, which is then normalized to obtain the Joint-Degree (JD). Combining DD and JD produces the spatio-temporal composite weight, which, together with the node key features, locates the key motion regions that determine the fundamental properties of actions. In the DGA module, based on the key motion regions located by DW-KRM, style transfer is first applied to nodes with lower JD values. Then, from both similarity and dissimilarity perspectives, different levels of data augmentation are applied to the key and non-key motion regions respectively, enriching the diversity of samples while maintaining semantic integrity. Finally, in the EDGRQ module, an entropy-based DG is introduced, where entropy serves as the utility function and dominance strategy acts as the convergence mechanism to enhance the value of negative samples. The overall framework of DoGCLR is illustrated in Fig.~\ref{fig: 01}, in which the positive sample optimization is achieved by DW-KRM and DGA, and the negative sample screening is performed by EDGRQ. Together, these components constitute the dual-dimensional dominance system of “positive sample optimization and negative sample selection”.

\subsection{Spatio-temporal Dual-Weight synergistic localization algorithm-based Key Region partitioning Module}

\begin{figure*}
  \begin{center}
  \includegraphics[width=0.7\textwidth]{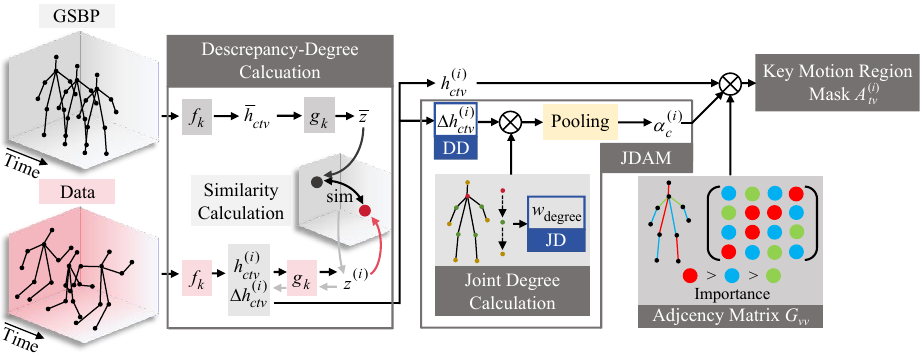}\\
  \caption{Block diagram of DW-KRM. The data and Global Statistical Benchmark Pose (GSBP) are encoded by the key encoder $f_k(\cdot)$ and MLP $g_k(\cdot)$ to compute their respective features. The Discrepancy-Degree (DD) is then fed into the Joint-Degree Activation Module (JDAM), where it is multiplied by the Joint-Degree (JD) calculated within JDAM to produce the spatio-temporal composite weight $\alpha_{c}^{(i)}$. This output from JDAM provides guidance for subsequent partitioned data augmentation. $\bigotimes$ denotes matrices multiplication.}\label{fig: 02}
  \end{center}
\end{figure*}

In self-supervised contrastive learning, positive samples are required to possess high information content and strong representativeness. High information content requires positive samples to contain rich fine-grained motion information, while strong representativeness requires positive samples to accurately reflect the core features of action categories based on the key joints of the skeleton. Therefore, this paper proposes a \textit{Spatio-temporal Dual-Weight synergistic localization algorithm}. After the sample is input into DW-KRM, based on the idea of local dominance, two types of weights — the Discrepancy-Degree (DD) and the Joint-Degree (JD) calculated in the Joint-Degree Activation Module (JDAM) — are fused to generate a key motion region mask, thereby achieving adaptive focus on key motion regions. The block diagram of DW-KRM is shown in Fig.~\ref{fig: 02}.

To facilitate the explanation of the proposed Spatio-temporal Dual-Weight synergistic localization algorithm, the definitions of \textit{Global Statistical Benchmark Pose (GSBP)}, \textit{Discrepancy-Degree (DD)}, and \textit{Joint-Degree (JD)} are introduced as follows.

\begin{definition}[Global Statistical Benchmark Pose, GSBP]
Let there be a skeleton dataset $\{X^{(1)}, X^{(2)}, \cdots, X^{(N)}\}$, where the total number of samples is $N$, and $X^{(i)} \in \mathbb{R}^{|C_X|\times|T|\times|V|}$, where $|C_X|$, $|T|$, and $|V|$ respectively denote the channel, frame, and joint set sizes of $X^{(i)}$. Then the Global Statistical Benchmark Pose (GSBP) is the algebraic mean of the dataset along the sample dimension, that is:

\begin{equation}
\overline{X} = \frac{1}{N}\sum_{i=1}^{N} X^{(i)}.
\label{eq: gsbp}\end{equation}
\end{definition}

\begin{definition}[Discrepancy-Degree, DD]
Let $X^{(i)}$ and $\overline{X}$ be the sample and the GSBP, respectively, whose features obtained through the key encoder $f_k(\cdot)$ and MLP $g_k(\cdot)$ are denoted as $z^{(i)}$ and $\overline{z}$. Their cosine similarity is defined as:
\begin{equation}
sim(z^{(i)}, \overline{z}) = \frac{z^{(i)} \cdot \overline{z}}{\|z^{(i)}\| \cdot \|\overline{z}\|},
\label{eq: sim}\end{equation}
where $\|\cdot\|$ denotes the $\ell_2$-norm of the feature vector. The Discrepancy-Degree (DD) of the $(t,v)$-th node of the $i$-th sample is defined as the negative derivative of the similarity $sim(z^{(i)}, \overline{z})$ with respect to the key feature $h^{(i)}_{ctv}$ on the channel of the encoded sample $X^{(i)}$:
\begin{equation}
\Delta h^{(i)}_{ctv} = -\frac{\partial sim(z^{(i)}, \overline{z})}{\partial h^{(i)}_{ctv}},
\label{eq: dd}\end{equation}
where $\Delta h^{(i)}_{ctv} \in \mathbb{R}^{|C|\times|T|\times|V|}$, and $|C|$, $|T|$, $|V|$ respectively denote the channel, frame, and joint set sizes of $\Delta h^{(i)}_{ctv}$, in which $|T|$, $|V|$ are completely consistent with those of $X^{(i)}$. $\Delta h^{(i)}_{ctv}$ denotes the element at position $(c,t,v)$ of $\Delta h^{(i)}$.
\end{definition}

\begin{definition}[Joint-Degree, JD]
Let the skeleton undirected graph of the $t$-th frame of the $i$-th sample be $G_t^{(i)}=(V_t^{(i)},E_t^{(i)})$, where $V_t^{(i)}$ denotes the node set and $E_t^{(i)}$ denotes the total number of nodes. The corresponding adjacency matrix is $A_t^{(i)}=(a_{vw})_{|V_t^{(i)}|\times|V_t^{(i)}|}$, and $|V_t^{(i)}|$ denotes the total number of nodes, $v,w\in V_t^{(i)}$. Then, the degree of node $v$ is defined as:
\begin{equation}
deg(v)=\sum_{w\in V_t^{(i)},\,w\neq v}a_{vw},
\label{eq: deg}\end{equation}
where $a_{vw}=1$ if there exists an edge between $v$ and $w$, otherwise $a_{vw}=0$. The Joint-Degree (JD) is the normalized value of the node degree, defined as:
\begin{equation}
w_{degree}[v]=\frac{|V_t^{(i)}| \cdot deg(v)}{\sum_{v\in V_t^{(i)}}deg(v)}.
\label{eq: jd}\end{equation}
\end{definition}

The three definitions above form the foundation of the Spatio-temporal Dual-Weight synergistic localization algorithm and provide both temporal and spatial constraints for subsequent region localization. First, the sample $X^{(i)}$ and the GSBP $\overline{X}$ are input into the key encoder and the MLP to obtain feature $z^{(i)}$ and $\overline{z}$. The DD is then computed by Eq.~(\ref{eq: dd}). Next, the calculated DD is input into JDAM, and the degree-based weight vector $w_{degree}$ is computed according to Eq.~(\ref{eq: jd}). By combining DD and JD, the spatio-temporal composite weight $\alpha_c^{(i)}$ is defined to measure the importance of neuron $c$, as shown in:
\begin{equation}
\alpha_c^{(i)}=\frac{1}{|T|\times|V|}\sum_{t\in T}\sum_{v\in V}\delta(\Delta h_{ctv}^{(i)})w_{degree}[v],
\label{eq: alpha}\end{equation}
where $|T|$, $|V|$ respectively denote the frame and joint set sizes of $\Delta h_{ctv}^{(i)}$, and $\delta(\cdot)$ denotes the activation function.  

Finally, the weighted combination of the composite weight and the features is calculated, and after activation, it is multiplied by the importance adjacency matrix to generate the key motion region mask:
\begin{equation}
A_{tv}^{(i)}=\delta\!\left(\sum_{c\in C}\alpha_c^{(i)}h_{ctv}^{(i)}\right)G_{vv},
\label{eq: atv}\end{equation}
where $G_{vv}$ is the adjacency matrix of the skeleton data used for importance smoothing. Through this linear combination, features that have a negative impact on similarity are selected. The key motion region corresponds to the region whose importance exceeds the threshold, while the remaining regions are considered non-key regions.  

From Eq.~(\ref{eq: alpha}), the spatio-temporal composite weight $\alpha_c^{(i)}$ considers both DD ($\Delta h_{ctv}^{(i)}$) and JD ($w_{degree}[v]$), exploring nodes with higher DD from the temporal perspective and focusing on nodes with higher JD in the spatial topology of the skeleton graph. Thus, the distribution of weights across key regions is determined collaboratively by the two factors DD and JD.

From the perspective of game theory, DW-KRM can be regarded as a local DG. In this game, each skeleton node acts as a player whose strategy is how to allocate attention weights between DD ($\Delta h_{ctv}^{(i)}$) in the temporal dimension and JD ($w_{degree}[v]$) in the spatial dimension to maximize its information gain in the global representation. The spatio-temporal composite weight $\alpha_c^{(i)}$ of a node can be regarded as its payoff function $\alpha_c^{(i)} = f(\Delta h_{ctv}^{(i)}, w_{degree}[v])$, and the competition and cooperation among all nodes over weight allocation form a non-cooperative DG system. Finally, the distribution of node weights converges to a local dominant equilibrium, meaning that no node can achieve higher payoff by adjusting its own strategy while others remain unchanged. This game-theoretic feature selection mechanism enables DW-KRM to possess adaptability and robustness in key-region extraction, thereby supporting the subsequent DGA.

The \textit{Spatio-temporal Dual-Weight synergistic localization algorithm} is shown in Algorithm~\ref{alg:st_dw_sla}.

\begin{figure*}
  \begin{center}
  \includegraphics[width=0.7\textwidth]{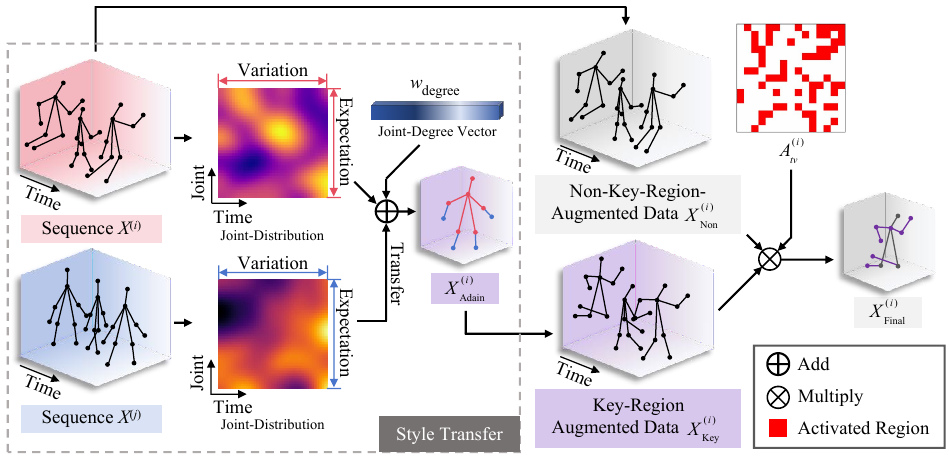}\\
  \caption{Block diagram of DGA. DGA first preserves the content of nodes in $X^{(i)}$ with higher Joint-Degree (JD) — that is, the spatial expectation and temporal variance of $X^{(i)}$ — and transfers the style (spatial expectation and temporal variance) of another sample $X^{(j)}$ to the nodes in $X^{(i)}$ with lower JD. Subsequently, according to the key motion region mask $A_{tv}^{(i)}$ output by DW-KRM, strong augmentations are applied to the key motion regions, while normal augmentations are applied to the non-key regions, thereby generating the partitioned augmented sequence $X_{Final}^{(i)}$.}\label{fig: 03}
  \end{center}
\end{figure*}

\begin{algorithm}[t]
\caption{The Spatio-temporal Dual-Weight synergistic localization algorithm.}
\label{alg:st_dw_sla}
\begin{algorithmic}[1]
\State $\mu \leftarrow$ mean\_over\_dataset($D$)
\Statex \hspace{0.0em}\texttt{\# GSBP}

\For{each sequence $x \in X$}
    \State Discrepancy representation $(x$ vs. $\mu)$, obtain similarity $s$ and gradient $g$
    \Statex \hspace{1.4em}\texttt{\# Discrepancy representation}
    
    \State $z_x = g_k(f_k(x));$ \quad $z_\mu = g_k(f_k(\mu))$
    \State $s = cosine(z_x, z_\mu)$
    \State $g = backprop\_gradient(s~w.r.t.~feature(x))$
    \Statex \hspace{1.4em}\texttt{\# compute DD}

    \State $w_{deg} = normalize(deg)$
    \Statex \hspace{1.4em}\texttt{\# compute JD}

    \State $A = GAP\_time\_joint(g \odot w_{deg})$
    \Statex \hspace{1.4em}\texttt{\# compute composite weight}

    \State $M = threshold(smooth(A; adjacency))$
    \Statex \hspace{1.4em}\texttt{\# compute mask matrix}

    \State store $M$ for $x$
\EndFor
\end{algorithmic}
\end{algorithm}

\subsection{Dual-scale-Game-based partitioned data Augmentation module}

Data augmentation plays a vital role in semantic feature extraction and generalization in contrastive learning. How to design more diverse transformations while maintaining task-relevant information remains a major challenge. Overly simple augmentation prevents the model from learning richer variations, while excessively difficult augmentation may destroy semantic integrity. To address this, we propose DGA, which first transfers the style of another randomly sampled instance $X^{(j)}$ onto the current instance $X^{(i)}$, and then, based on the key-region localization results from DW-KRM, constructs a dual-scale Dominance Game (DG). This enables finer-grained augmentations that explore richer motion patterns while preserving sample semantics. The block digram of DGA is shown in Fig.~\ref{fig: 03}.

First, to enhance both semantic consistency and diversity, the style of sample $X^{(j)}$ is transferred into sample $X^{(i)}$. 
Nodes with higher JD are regarded as content nodes, focusing on dynamic information strongly correlated with the action category; 
nodes with lower JD are treated as style nodes, corresponding to global structure irrelevant to action semantics. A different sample $X^{(j)}$ is randomly selected from the dataset, and the spatial expectation $\mu(X^{(i)}), \mu(X^{(j)})$ and temporal variance $\sigma^2(X^{(i)}), \sigma^2(X^{(j)})$ are computed respectively. Then style transfer is performed as follows:
\begin{equation}
\begin{aligned}
X^{(i)}_{Adain} &= w_{degree}[v] \cdot X^{(i)} + (1 - w_{degree}[v]) \\
&\quad \times\left(\frac{X^{(j)} - \mu(X^{(i)})}{\sigma(X^{(i)})}\cdot \sigma(X^{(j)}) + \mu(X^{(j)})\right),
\end{aligned}
\label{eq: style}\end{equation}
where $X^{(i)}_{Adain}$ is the style-transferred sample, $w_{degree}[v]$ is the JD of node $v$, and $\mu(\cdot)$, $\sigma^2(\cdot)$ denote spatial expectation and temporal variance respectively.

After style transfer, although semantic consistency is maintained, the strategy for optimizing motion patterns remains lacking. 
Therefore, a dual-scale dominance-game mechanism is introduced to adaptively select augmentation strategies. Let the game $\Gamma_1 = (\{1\}, C_1, u_1)$ represent a single-player game in which the player is the key-region augmentation module, and $C_1$ contains various augmentation operations. Eight strong augmentation strategies $c_{\mathcal{T'}}$ and two normal ones $c_{\mathcal{T}}$ are designed:

(1) \textbf{Strong augmentations $c_{\mathcal{T'}}$}: four spatial transformations (shearing, flipping, rotation, and axis masking), two temporal transformations (cropping, temporal flipping), and two spatiotemporal ones (Gaussian noise, Gaussian blur).

(2) \textbf{Normal augmentation $c_{\mathcal{T}}$}: adding high-variance Gaussian noise and skeleton mixing.

The objective of the game is to maximize dissimilarity between the original and augmented samples in key regions (to explore diverse motion modes) while maintaining similarity in non-key regions (to preserve semantics). The payoff function is defined as:
\begin{equation}
\begin{aligned}
u_1(c) &= \sum_{v \in V} \sum_{t \in T} 
\Big(M_{tv}^{(i)} \cdot Dissimilarity_{tv}(c) \\
&\quad + (1 - M_{tv}^{(i)}) \cdot Similarity_{tv}(c)\Big),
\end{aligned}
\label{eq: u1}
\end{equation}
where $M_{tv}^{(i)} = 1$ if $A_{tv}^{(i)}$ exceeds the importance threshold hyperparameter $\theta$, otherwise $M_{tv}^{(i)} = 0$. $Similarity_{tv}(c)$ and $Dissimilarity_{tv}(c)$ denote the similarity and dissimilarity scores of node $(t,v)$ under strategy profile $c$. It is defined that: $Similarity_{tv}(c_{\mathcal{T}'}) = 1$, $Similarity_{tv}(c_{\mathcal{T}}) = 2$, and $Dissimilarity_{tv}(c) = 3 - Similarity_{tv}(c)$. According to Eq.~(\ref{eq: u1}), when $M_{tv}^{(i)} = 1$, let $c_{tv} = c_{\mathcal{T}'}$; when $M_{tv}^{(i)} = 0$, let $c_{tv} = c_{\mathcal{T}}$. Thereby $u_1(c)$ can always achieve the maximum value strictly. Therefore, the strategy profile $(c_{Key}, c_{Non}) = (c_{\mathcal{T}'}, c_{\mathcal{T}})$ constitutes a dominance strategy of game $\Gamma_1$, where $c_{Key}$ denotes the strategy adopted by nodes in the key region, and $c_{Non}$ denotes the strategy adopted by nodes in the non-key region.

Finally, according to the dominance equilibrium of the game $(c_{Key}, c_{Non}) = (c_{\mathcal{T}'}, c_{\mathcal{T}})$, the dual-scale dominance-based augmentation strategy is formulated. For regions where $M_{tv}^{(i)} = 1$, apply strong augmentation to the style-transferred sample $X_{Adain}^{(i)}$ to obtain $X_{Key}^{(i)}$; for regions where $M_{tv}^{(i)} = 0$, apply normal augmentation to $X^{(i)}$ to obtain $X_{Non}^{(i)}$. The two are then combined as:
\begin{equation}
X_{Final}^{(i)} = M^{(i)} \odot X_{Key}^{(i)} + (1 - M^{(i)}) \odot X_{Non}^{(i)},
\label{eq: final}
\end{equation}
where $X_{Final}^{(i)} \in \mathbb{R}^{|T| \times |V|}$ denotes the final augmented data. $M^{(i)} = (M_{tv}^{(i)})_{|T| \times |V|} \in \mathbb{R}^{|T| \times |V|}$ represents a binary-encoded matrix. For skeleton $i$, if the key motion region mask $A_{tv}^{(i)}$ of joint $v$ in frame $t$ exceeds the importance threshold $\theta$, then the element $M_{tv}^{(i)}$ is set to $1$; otherwise $M_{tv}^{(i)}$ is set to $0$. The operator $\odot$ denotes the Hadamard product, i.e., element-wise multiplication of matrices.

DGA through style transfer enhances the semantic diversity and rationality of sample $X^{(i)}$. By applying different data augmentation strategies to different regions, DGA preserves the semantics of $X^{(i)}$ while exploring a broader range of behavioral patterns. Meanwhile, since the game $\Gamma_1$ possesses a unique dominance strategy $(c_{Key}, c_{Non}) = (c_{\mathcal{T}'}, c_{\mathcal{T}})$, $\Gamma_1$ is a Dominance Game (DG). Furthermore, as long as $|Similarity_{tv}(c) - Similarity\_True_{tv}(c)| < 0.5$, the strategy profile $(c_{Key}, c_{Non}) = (c_{\mathcal{T}'}, c_{\mathcal{T}})$ consistently remains the dominance strategy, demonstrating the strong robustness of the dominance-strategy mechanism proposed in this paper.

\subsection{Entropy-driven Dominance Game Replacement Queue}

In contrastive learning, traditional methods often adopt a first-in-first-out (FIFO) updating strategy, treating all negative samples equally. However, this approach fails to exploit the differences among negative samples, particularly those that are more challenging. After completing the local and global augmentation of positive samples, the network needs to maintain a dynamically balanced information distribution during negative sample construction. Therefore, this paper introduces an Entropy-driven Dominance Game Replacement Queue (EDGRQ), which optimizes the quality of negative samples from the perspective of sample strategy, ensuring that DoGCLR always trains with the most informative and effective negative samples in each iteration.

\begin{algorithm}[t]
\caption{DoGCLR algorithm.}
\label{alg:dogclr}
\begin{algorithmic}[1]

\State \textbf{Input:} 
\Statex \makebox[3.7em][l]{$D$}          \texttt{\# skeleton dataset}
\Statex \makebox[3.7em][l]{$(f_q, f_k)$} \texttt{\# encoders}
\Statex \makebox[3.7em][l]{$(g_q, g_k)$} \texttt{\# projection heads}
\Statex \makebox[3.7em][l]{$M$}          \texttt{\# EDGRQ}
\Statex \makebox[3.7em][l]{$\tau$}          \texttt{\# temperature}
\Statex \makebox[3.7em][l]{$m$}          \texttt{\# momentum}
\Statex \makebox[3.7em][l]{$(\mathcal{T}_s, \mathcal{T}_n)$} \texttt{\# augmentations}

\vspace{0.5em}
\State \textbf{Procedure:}
\State Initialize encoders and projection heads; copy parameters to momentum teacher branch
\Statex \hspace{0.0em}\texttt{\# initialization}
\State Initialize memory bank $Q \leftarrow \varnothing$
\Statex \hspace{0.0em}\texttt{\# initialize memory bank}

\For{each minibatch $\mathcal{X} \subset D$}
    \State Generate key region masks using DW-KRM + JDAM
    \Statex \hspace{1.4em}\texttt{\# key-region mask generation}
    
    \State Construct two augmented views with DGA
    \Statex \hspace{1.4em}\texttt{\# dual-scale data augmentation}
    
    \State Encode and project to obtain queries $q$ and keys $k$
    \Statex \hspace{1.4em}\texttt{\# feature encoding and projection}
    
    \State Compute InfoNCE loss with positive pairs $(q_i, k_i)$ and negatives from $M$
    \Statex \hspace{1.4em}\texttt{\# contrastive loss computation}
    
    \State Update $(f_q, g_q)$ by gradient descent; momentum update $(f_k, g_k)$
    \Statex \hspace{1.4em}\texttt{\# momentum encoder update}
    
    \State Update memory bank $M$ using EDGRQ: retain high-entropy samples, discard low-entropy ones
    \Statex \hspace{1.4em}\texttt{\# entropy-driven queue replacement}
\EndFor

\vspace{0.3em}
\State \textbf{Output:} pretrained encoder $f_q$ (or $f_k$) for downstream evaluation
\Statex \hspace{0.0em}\texttt{\# final model for evaluation}

\end{algorithmic}
\end{algorithm}

A Dominance Game (DG) is defined as $\Gamma_2 = (\{1\}, C_2, u_2)$: the player set $N = \{1\}$ indicates that this game has only one player — the negative sample memory bank. The strategy set $C_2 = \{c_1, c_2, \cdots, c_{|M|}\}$, where $|M|$ denotes the size of the memory bank, and $c_i$ $(i = 1,2,\cdots,|M|)$ represents the strategy ``replacing the $i$-th sample in the memory bank with the current sample''. The payoff function of the game adopts the information entropy function, as shown in Eq.~(\ref{eq:entropy_payoff}):
\begin{equation}
u_2(c_i) = - \sum_{j=1}^{|M|} p_j(c_i) \log p_j(c_i), \quad i = 1,2,\cdots,|M|,
\label{eq:entropy_payoff}
\end{equation}
where $p_j(c_i)$ denotes the normalized similarity probability, computed as Eq.~(\ref{eq:similarity_prob}):
\begin{equation}
p_j(c_i) = \frac{\exp(sim(s^{(l+1)}, m_j))}{\sum_{k=1}^{|M|} \exp(sim(s^{(l+1)}, m_k))},\quad j = 1,2,\cdots,|M|,
\label{eq:similarity_prob}
\end{equation}
where $sim(\cdot, \cdot)$ denotes the cosine similarity between two samples; $m_j$, $m_k$ are the negative sample features in the memory bank after replacing sample $m_i$ with the positive sample $s^{(l)}$; $s^{(l)}$ and $s^{(l+1)}$ represent the positive sample features at the $l$-th and $(l+1)$-th training iterations, respectively.

According to the functional property of $u_2(c)$, when
\begin{equation}
c_n \in \mathop{\arg\max}_{c_i \in C_2}u_2(c_i),
\label{eq:dominance_strategy}
\end{equation}
i.e., $p_j(c_n) = 1 / |M|$, for any $j = 1,2,\cdots,|M|$, the payoff $u_2(c_n)$ achieves the global optimum, and the game converges to the dominance strategy at this moment.

Consequently, during the $(l+1)$-th training iteration, the negative samples in the memory bank always exhibit maximum uncertainty with respect to $s^{(l+1)}$. This avoids the quality fluctuation caused by indiscriminate processing under the FIFO strategy and maximizes the utilization efficiency of high-value negative samples. As a result, the network continually focuses on key discriminative tasks during contrastive learning, thereby significantly enhancing the discriminability and generalization of its feature representations.

\subsection{Pseudocode of the Algorithm}

The pseudocode of the proposed \textit{DoGCLR network} is shown in Algorithm~\ref{alg:dogclr}.

\section{Experiments}

In Section A, we introduce the datasets used in our experiments. Section B presents the experimental setup, followed by the performance evaluation and ablation studies in Section C and D, respectively.

\subsection{Datasets}

The experiments in this paper are conducted on two publicly available human action datasets: NTU RGB+D \cite{Amir2016NTU60} and PKU-MMD \cite{Liu2017PKUMMD}, to evaluate the feasibility and effectiveness of the proposed methods.


\subsubsection{NTU RGB+D} 
The NTU RGB+D dataset, released by Nanyang Technological University (NTU) in 2016, is a large-scale benchmark for human action recognition using RGB+D data. It contains 60 action classes and 56,578 sequences recorded by three Microsoft Kinect v2 cameras, including RGB videos, depth maps, skeleton data, and infrared sequences. Data were collected from 40 subjects under three camera views. Two protocols are provided: Cross-Subject (X-Sub), where subjects 1–20 are used for training and 21–40 for testing, and Cross-View (X-View), where two views are used for training and one for testing. An extended version, NTU RGB+D 120, was released in 2019, increasing the number of classes to 120 and the total samples to about 114,480. It involves 106 subjects and 32 camera setups, offering greater scene diversity and complexity. Similar to NTU60, it adopts Cross-Subject (X-Sub) and Cross-Setup (X-Set) protocols, making it one of the most challenging skeleton-based benchmarks.


\subsubsection{PKU-MMD} 
The PKU-MMD dataset, released by the Wangxuan Institute of Peking University in 2017, provides a large-scale multimodal benchmark for human action recognition. It consists of two subsets, Part I and Part II. Part I includes 66 subjects performing 51 actions, producing 20,734 samples and 1,074 independent action instances. Under the C-view protocol, data from the middle and right cameras are used for training, and those from the left camera for testing. Part II contains more complex and dynamic motion scenes with stronger inter-person interactions and occlusions. It follows the same C-view protocol but focuses on large-scale movements and real-world complexity, making it a valuable dataset for evaluating model robustness and generalization in challenging environments.

\subsection{Experimental Setup}

The detailed system configuration used in this paper is presented in Table~\ref{tab:sysconfig}.

\begin{table}[t]
\centering
\caption{System Configuration}
\label{tab:sysconfig}
\renewcommand{\arraystretch}{1.2}
\begin{tabular}{|c c|}
\hline
System Configuration & Parameters \\ \hline
CUDA Version & Cuda release V11.8.89 \\
Pytorch Version & Pytorch 1.13.0+cu117 \\
Python Version & 3.9.12 \\
GPU Model & Nvidia GeForce RTX3090 (24G) \\ \hline
\end{tabular}
\end{table}

In addition, this paper is built upon MoCov2~\cite{He2019MoCo} pretraining framework and employs an online (query) + offline (key) dual-encoder architecture for feature extraction. The overall network optimization adopts a InfoNCE loss function~\cite{He2019MoCo, Guo2022AimCLR, Lin2023ActionletDependent}. During training, the query encoder is updated via the backpropagation of the loss function, while the key encoder is updated through a momentum-based update mechanism, with the momentum coefficient set to 0.99. Experiments are conducted on the NTU RGB+D and PKU-MMD datasets to evaluate the performance of the proposed methods. The evaluation is performed from two perspectives: (1) Accuracy, to quantitatively compare the proposed method with existing self-supervised learning approaches; and (2) Visualization, to qualitatively analyze the learned representations.

\subsection{Performance Evaluation}

\begin{figure*}[!t]
    \centering
    \includegraphics[width=\textwidth]{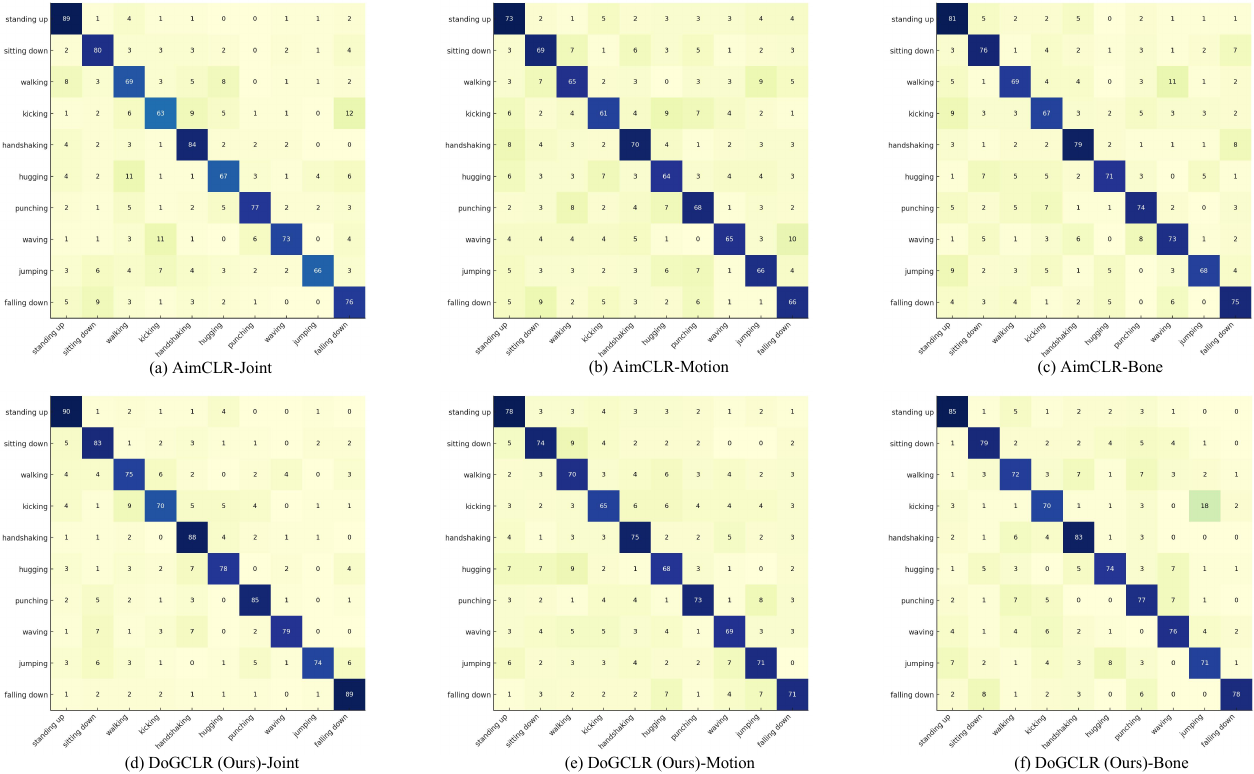}
    \caption{Confusion matrices of AimCLR and DoGCLR under the linear evaluation protocol on the NTU RGB+D 60 dataset, visualized for the Joint, Motion, and Bone streams. Each matrix is computed from 100 samples per class across 10 actions. DoGCLR consistently demonstrates higher recognition accuracy and reduced misclassification.}
    \label{fig:confusion_joint_motion_bone}
\end{figure*}

\begin{table*}[!t]
\centering
\caption{Comparison Results of the Proposed DoGCLR and Baseline Methods on NTU RGB+D 60, NTU RGB+D 120, and PKU-MMD Datasets under the Linear Evaluation Protocol}
\label{tab:dogclr_comparison}
\renewcommand{\arraystretch}{1.15}
\begin{tabular}{|ccccccc|}
\hline
\multirow{2}{*}{Method} & 
\multirow{2}{*}{Stream} & 
\multicolumn{2}{c}{NTU RGB+D 60 (\%)} & 
\multicolumn{2}{c}{NTU RGB+D 120 (\%)} & 
\multicolumn{1}{c|}{PKU-MMD (\%)} \\ \cline{3-7}
 & & X-Sub Acc.$\uparrow$ & X-View Acc.$\uparrow$ & X-Sub Acc.$\uparrow$ & X-Set Acc.$\uparrow$ & Part I Acc.$\uparrow$ \\ \hline

SkeletonCLR~\cite{Li2021CrossViewConsistency} & Joint & 68.3 & 76.4 & 56.8 & 55.9 & 80.9 \\
AimCLR~\cite{Guo2022AimCLR} & Joint & \underline{74.3} & \underline{79.6} & \underline{63.4} & \underline{63.4} & \underline{83.0} \\
\textbf{DoGCLR (Ours)} & Joint & \textbf{76.7+2.4} & \textbf{83.3+3.7} & \textbf{67.7+4.3} & \textbf{71.4+8.0} & \textbf{83.2+0.2} \\ \hline

SkeletonCLR~\cite{Li2021CrossViewConsistency} & Motion & 53.3 & 50.8 & 39.6 & 40.2 & 63.4 \\
AimCLR~\cite{Guo2022AimCLR} & Motion & \underline{66.8} & \underline{70.6} & \underline{57.3} & \underline{54.4} & \underline{72.0} \\
\textbf{DoGCLR (Ours)}  & Motion & \textbf{71.4+4.6} & \textbf{77.2+6.6} & \textbf{61.3+4.0} & \textbf{59.7+5.3} & \textbf{77.4+5.4} \\ \hline

SkeletonCLR~\cite{Li2021CrossViewConsistency} & Bone & 69.4 & 67.4 & 48.4 & 52.0 & 72.6 \\
AimCLR~\cite{Guo2022AimCLR} & Bone & \underline{73.2} & \underline{77.0} & \underline{62.9} & \underline{63.4} & \underline{82.0} \\
\textbf{DoGCLR (Ours)}  & Bone & \textbf{76.5+3.3} & \textbf{80.0+3.0} & \textbf{64.9+2.0} & \textbf{65.8+2.4} & \textbf{87.1+5.1} \\ \hline

3s-SkeletonCLR~\cite{Li2021CrossViewConsistency} & 3s & 75.0 & 79.8 & 60.7 & 62.6 & 85.3 \\
3s-AimCLR~\cite{Guo2022AimCLR} & 3s & \underline{78.2} & \underline{83.8} & \underline{68.2} & \underline{68.8} & \underline{87.8} \\
\textbf{3s-DoGCLR (Ours)}  & 3s & \textbf{81.1+2.9} & \textbf{89.4+5.6} & \textbf{71.2+3.0} & \textbf{75.5+6.7} & \textbf{90.0+2.2} \\ \hline

\end{tabular}
\end{table*}

To validate the effectiveness of the proposed network, the DoGCLR algorithm is compared with several recent state-of-the-art self-supervised learning methods. Following the evaluation protocol, two types of downstream tasks are adopted: the linear evaluation protocol and the K-nearest neighbor (KNN) evaluation protocol, which assess performance from both qualitative and quantitative perspectives. For each downstream task, the network parameters are obtained after pretraining under the self-supervised contrastive learning framework.


\subsubsection{Linear Evaluation Protocol}

In the linear evaluation setting, the pretrained action encoder is frozen, and a fully connected layer with a softmax function is appended and trained in a supervised manner. The linear evaluation is performed with a learning rate of 0.1 for a total of 100 epochs.

\begin{figure}[t]
    \centering
    \includegraphics[width=\columnwidth]{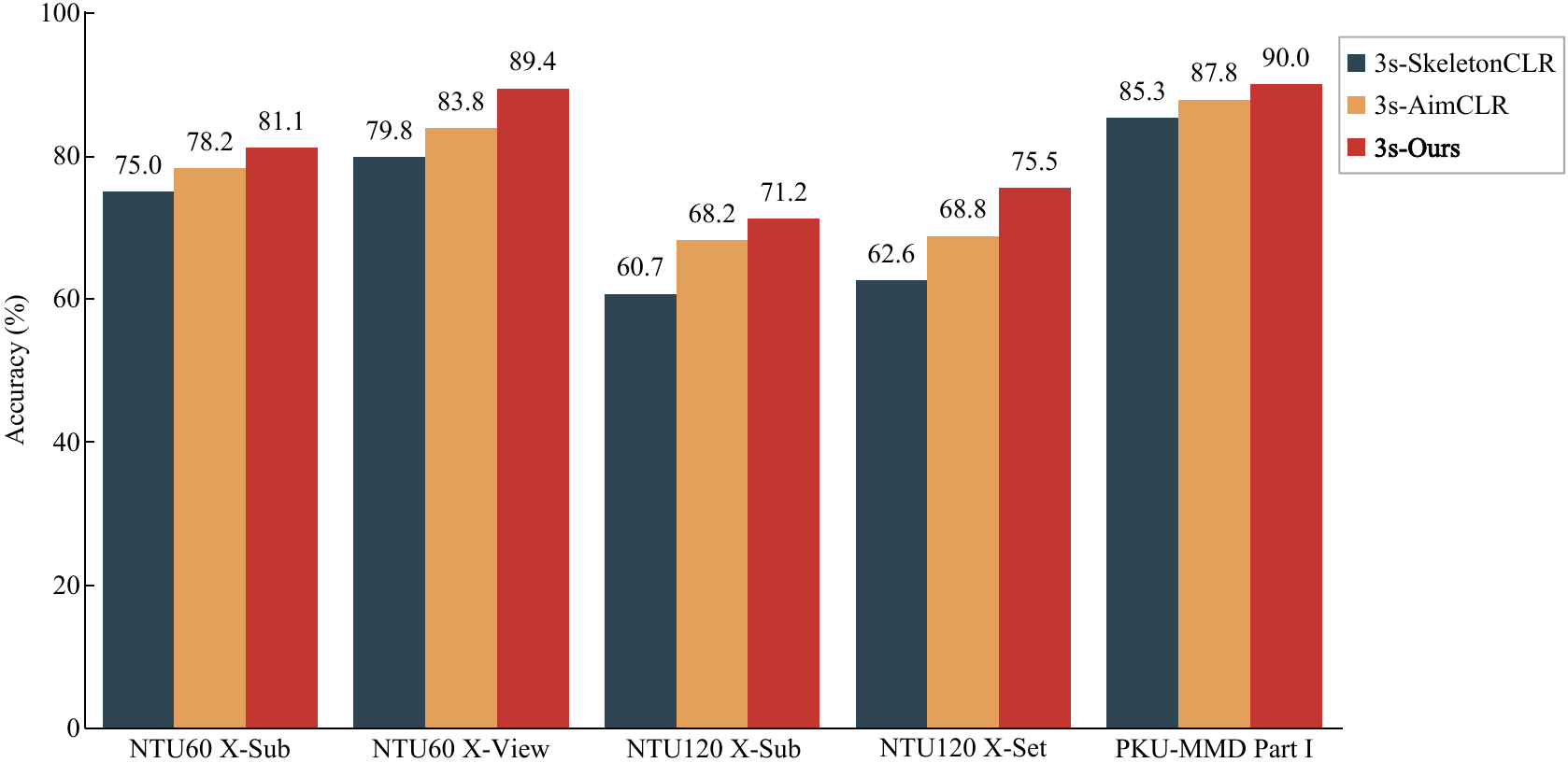}
    \caption{Comparison results of the 3s-baseline methods and 3s-DoGCLR on NTU RGB+D 60, NTU RGB+D 120, and PKU-MMD Part I datasets under the linear evaluation protocol. 3s-DoGCLR achieves consistently superior performance across all methods.}
    \label{fig:bar_3s_results}
\end{figure}

\textbf{Comparison with Baseline Methods.}  
Under the linear evaluation protocol, DoGCLR is compared with the baseline methods SkeletonCLR and AimCLR across multiple datasets and modalities. The results show consistent improvements. On the NTU RGB+D 60 dataset, DoGCLR achieves 81.1\% (X-Sub) and 89.4\% (X-View) in the three-stream setting, exceeding AimCLR by 2.9\% and 5.6\%, respectively. On NTU RGB+D 120, it reaches 71.2\% (X-Sub) and 75.5\% (X-Set), surpassing AimCLR by 3.0\% and 6.7\%. On PKU-MMD, DoGCLR attains 90.0\%, a 2.2\% improvement over AimCLR, demonstrating robustness on large-scale cross-scene data. Notably, on the Motion stream, the combination of DW-KRM and DGA yields over 5\% gain, enhancing motion discriminability and temporal modeling. 
Detailed comparisons are shown in Table~\ref{tab:dogclr_comparison}, where bold and underlined numbers denote the best and second-best results, respectively. The three-stream results on NTU60, NTU120, and PKU-MMD Part~I are illustrated in Fig.~\ref{fig:bar_3s_results}.

\begin{table*}[!t]
\centering
\caption{Comparison Results of the Proposed DoGCLR and Other Approaches on the NTU RGB+D 60 and NTU RGB+D 120 Datasets under the Linear Evaluation Protocol}
\label{tab:ntu}
\renewcommand{\arraystretch}{1.15}
\begin{tabular}{|c c c c c c|}
\hline
\multirow{2}{*}{Method} & 
\multirow{2}{*}{Backbone} & 
\multicolumn{2}{c}{NTU RGB+D 60 (\%)} & 
\multicolumn{2}{c|}{NTU RGB+D 120 (\%)} \\ \cline{3-6}
 & & X-Sub Acc.$\uparrow$ & X-View Acc.$\uparrow$ & X-Sub Acc.$\uparrow$ & X-Set Acc.$\uparrow$ \\ \hline
CorsSCLR (2021) \cite{Li2021CrossViewConsistency} & ST-GCN & 77.8 & 83.4 & 67.9 & 66.7 \\
MCAE (2021) \cite{Xu2021CapsuleAE} & MCAE & 65.6 & 74.7 & 52.8 & 54.7 \\
CP-STN (2021) \cite{Chen2025VLSkeleton} & ST-GCN & 69.4 & 76.6 & 55.7 & 54.7 \\
ST-CL (2022) \cite{Gao2023EfficientSTCL} & GCN & 68.1 & 69.4 & 54.2 & 55.6 \\
AimCLR (2022) \cite{Guo2022AimCLR} & ST-GCN & 78.9 & 83.8 & 68.2 & 68.8 \\
SkeAttnCLR (2023) \cite{Hua2023PartAwareCLR} & ST-GCN & 78.2 & 82.6 & 65.5 & 65.7 \\
ASAR (2023) \cite{Wang2023SS3D} & ST-GCN & 75.6 & 80.7 & 62.5 & 62.8 \\
HiCLR (2023) \cite{Dong2023HiCLR} & ST-GCN & 80.4 & 85.5 & 70.0 & 70.4 \\
ViA (2024) \cite{yang2024view} & GCN & 78.1 & 85.8 & 69.2 & 66.9 \\
CMCS (2024) \cite{liu2024cross} & ST-GCN & 78.6 & 84.5 & 68.5 & 71.1 \\
AimCLR++ (2024) \cite{Guo2024AimCLRPlus} & ST-GCN & 80.9 & 85.4 & \underline{70.1} & 71.2 \\
ISSM (2025) \cite{tu2025issm} & bi-GRU & \underline{81.0} & \underline{86.7} & 69.9 & \underline{73.2} \\
\textbf{DoGCLR (Ours)} & ST-GCN & \textbf{81.1} & \textbf{89.4} & \textbf{71.2} & \textbf{75.5} \\ \hline
\end{tabular}
\end{table*}

\begin{table*}[!t]
\centering
\caption{Comparison Results of the Proposed DoGCLR and Other Approaches on the PKU-MMD Part I and Part II Dataset under the Linear Evaluation Protocol }
\label{tab:pkummd}
\renewcommand{\arraystretch}{1.15}
\begin{tabular}{|c c c c|}
\hline
\multirow{2}{*}{Method} & 
\multirow{2}{*}{Backbone} & 
\multicolumn{2}{c|}{PKU-MMD (\%)} \\ \cline{3-4}
 &  & Part I Acc.$\uparrow$ & Part II Acc.$\uparrow$ \\ \hline
CorsSCLR (2021) \cite{Li2021CrossViewConsistency} & ST-GCN & 84.9 & 21.2 \\
AimCLR (2022) \cite{Guo2022AimCLR} & ST-GCN & 87.8 & 38.5 \\
SkeAttnCLR (2023) \cite{Hua2023PartAwareCLR} & ST-GCN & 83.8 & --- \\
ASAR (2023) \cite{Wang2023SS3D} & ST-GCN & 83.5 & 38.8 \\
CMCS (2024) \cite{liu2024cross} & ST-GCN & 88.1 & 39.6 \\
AimCLR++ (2024) \cite{Guo2024AimCLRPlus} & ST-GCN & \textbf{90.4} & \underline{41.2} \\ 
ISSM (2025) \cite{tu2025issm} & bi-GRU & 89.6 & 40.3 \\
\textbf{DoGCLR (Ours)} & ST-GCN & \underline{90.0} & \textbf{43.1} \\ \hline
\end{tabular}
\end{table*}

\begin{figure}[t]
    \centering
    \includegraphics[width=\columnwidth]{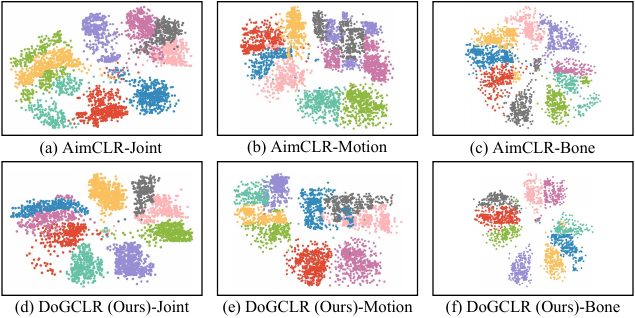}
    \caption{Feature distribution visualization of nine representative actions on the NTU RGB+D 60 dataset. The scatter maps show the encoded features after linear evaluation for AimCLR and DoGCLR, where DoGCLR exhibits more compact intra-class clustering and clearer inter-class separation.}
    \label{fig:feature_scatter_joint_motion_bone}
\end{figure}

To compare DoGCLR with AimCLR, we sample 100 instances from 10 representative actions — standing up, sitting down, walking, kicking, handshaking, hugging, punching, waving, jumping, and falling down — from the NTU RGB+D 60 dataset. As shown in Fig.~\ref{fig:confusion_joint_motion_bone}, DoGCLR achieves higher accuracy across all categories under linear evaluation for the Joint, Motion, and Bone streams. Furthermore, to visualize feature separability, 9 representative actions — standing up, sitting down, walking, handshaking, hugging, punching, waving, jumping, and falling down — are projected into a low-dimensional space, as shown in Fig.~\ref{fig:feature_scatter_joint_motion_bone}. DoGCLR presents tighter intra-class clusters and clearer inter-class boundaries, demonstrating stronger discriminative capability.

\textbf{Comparison with State-of-the-Art Methods.} 
Table~\ref{tab:ntu} presents the comparison results of the proposed method on the NTU RGB+D dataset (including NTU60 and NTU120) under the Cross-Subject (X-Sub) and Cross-View/Cross-Setup (X-View/X-Set) protocols. For NTU60, DoGCLR achieves 81.1\% (X-Sub) and 89.4\% (X-View),  outperforming SOTA method ISSM (81.0\%, 86.7\%) and for NTU120, DoGCLR achieves 71.2\% (X-Sub) and 75.5\% (X-Set), outperforming SOTA methods by 1.1\% and 2.3\% respectively. Compared with the enhanced variant AimCLR++ of the baseline method AimCLR, DoGCLR also shows strong competitiveness. For NTU60 X-Sub/X-View, DoGCLR surpasses AimCLR++ by 0.2\% and 4.0\% respectively. Furthermore, for NTU120, which involves more subjects, actions, and view variations, DoGCLR also surpasses AimCLR++ (70.1\%, 71.2\%). These results indicate that DoGCLR maintains excellent generalization on larger and more complex benchmarks, effectively capturing long-term temporal dependencies and Cross-View motion consistency, confirming the robustness of the proposed dominance-based contrastive framework. In particular, the improvements on NTU120 demonstrate that DoGCLR adapts well to data imbalance and intra-class diversity, showing stronger feature invariance and more stable convergence during linear evaluation.

A further comparison is conducted on the PKU-MMD Part I and Part II datasets, as summarized in Table~\ref{tab:pkummd}. 
The PKU-MMD dataset includes more realistic and fine-grained motion interactions, providing a complementary evaluation to the NTU benchmarks. Part I contains fewer samples and simpler motion patterns. Since the difference between the Global Statistic Benchmark Pose (GSBP) and the true GSBP depends on dataset size, a smaller dataset leads to reduced differentiation in Discrepancy-Degree (DD) among joints. Consequently, DoGCLR may show slight bias in localizing key regions, resulting in performance in Table~\ref{tab:pkummd} that is not the highest. However, its accuracy is only 0.4\% lower than the best algorithm and 7.0\% higher than the baseline AimCLR, while still outperforming most existing methods and maintaining stable recognition capability on small-scale data. This suggests that the proposed dominance-guided mechanism preserves its discriminative potential even when structural cues are limited.

On the more challenging PKU-MMD Part II dataset, which involves complex motions and intensive inter-person interactions, DoGCLR achieves 43.1\% accuracy, outperforming AimCLR (38.5\%) and AimCLR++ (41.2\%). This improvement benefits from DoGCLR’s precise localization of key motion regions and effective retention of hard negative samples, which enhance feature discrimination under high intra-class variation. Moreover, the cooperative modeling of Joint, Motion, and Bone streams enables DoGCLR to extract richer motion cues and structural correlations, improving robustness to temporal occlusion and view variation. These results further validate the effectiveness and scalability of the proposed dominance-based contrastive learning framework across datasets of varying scales, complexities, and motion patterns.


\subsubsection{K-Nearest-Neighbor (KNN) Evaluation Protocol}

\begin{figure}[t]
    \centering
    \includegraphics[width=0.5\textwidth]{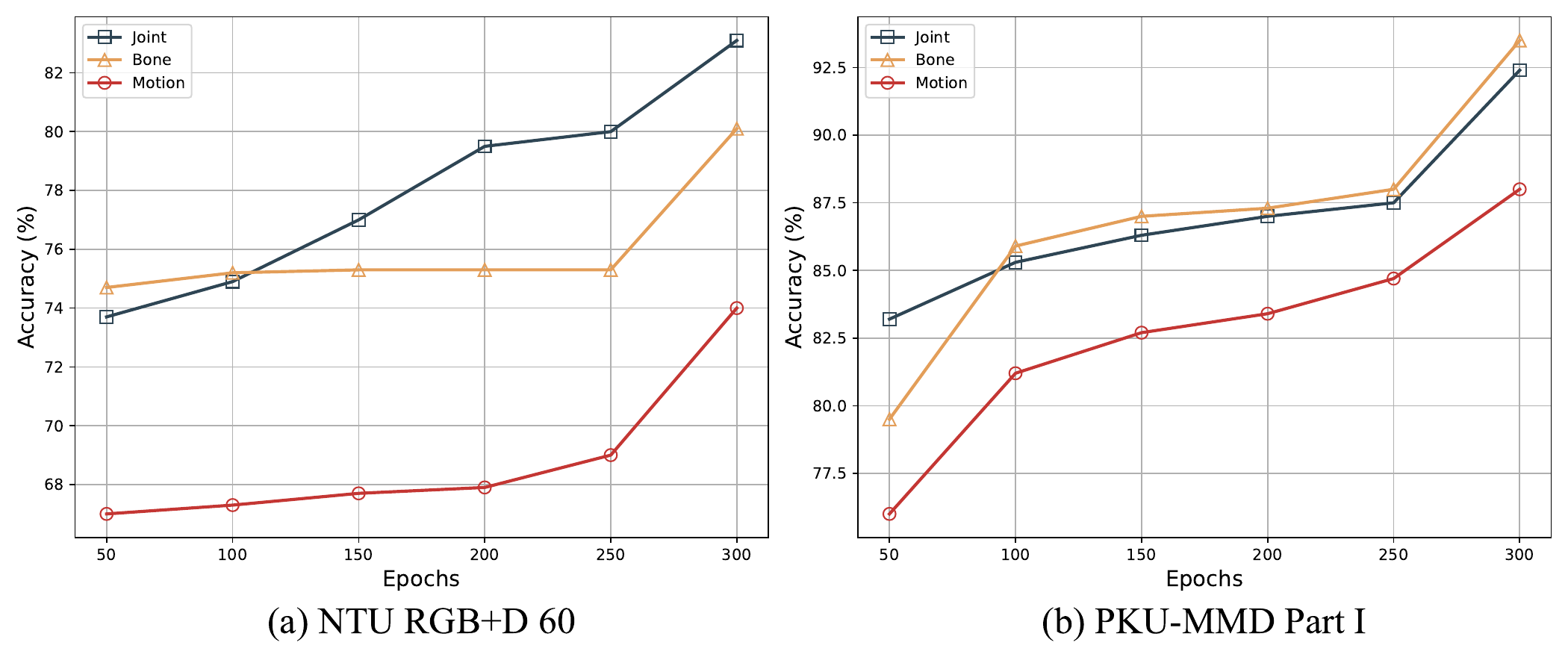}
    \caption{K-Nearest-Neighbor evaluation of the three-stream skeleton representation.}
    \label{fig: 04}
\end{figure}

The K-Nearest-Neighbor (KNN) evaluation is an instance-based learning approach. For a test sample, its similarity to each training sample is computed, and the K most similar samples (neighbors) are selected. The class of the test sample is then determined by a voting mechanism based on the labels of these nearest neighbors. To verify the effectiveness of the proposed method under the KNN evaluation protocol, experiments are conducted on both the NTU RGB+D 60 and PKU-MMD Part I datasets. The comparison focuses on the recognition accuracy of the proposed model using single-stream (skeleton-only) data. The experimental results are illustrated in Fig.~\ref{fig: 04}. When evaluated at the 300th epoch, both datasets show a significant improvement in accuracy compared with earlier epochs. It demonstrates that the proposed DoGCLR model possesses strong feature-discrimination capability during the early training stage, effectively capturing high-quality representations even before full convergence.

\subsection{Ablation Studies}

\subsubsection{Module-Wise Ablation Analysis}

\begin{table*}[!t]
\centering
\caption{Model-Wise Ablation Results of the Proposed Algorithm on the NTU RGB+D 60 Dataset (Linear Evaluation Protocol)}
\label{tab:ablation_ntu60}
\renewcommand{\arraystretch}{1.15}
\begin{tabular}{|c c c c c c|}
\hline
\multirow{2}{*}{Basic Augmentation} & 
\multirow{2}{*}{DW-KRM/JDAM} & 
\multirow{2}{*}{DGA} & 
\multirow{2}{*}{EDGRQ} & 
\multicolumn{2}{c|}{NTU RGB+D 60 (\%)} \\ \cline{5-6}
 &  &  &  & X-Sub Acc.$\uparrow$ & X-View Acc.$\uparrow$ \\ \hline
\checkmark &   &   &   & 71.1 & 79.5 \\
\checkmark & \checkmark &   &   & 73.2 & 77.1 \\
\checkmark & \checkmark & \checkmark &   & \underline{77.4} & \underline{82.8} \\
\textbf{\checkmark} & \textbf{\checkmark} & \textbf{\checkmark} & \textbf{\checkmark} & \textbf{81.1} & \textbf{89.4} \\ \hline
\end{tabular}
\end{table*}

To further verify the performance contribution of each component within the proposed framework, a progressive module-wise ablation experiment is conducted on the NTU RGB+D 60 dataset. This analysis evaluates the individual effects of the DGA, DW-KRM/JDAM, and EDGRQ modules on the overall recognition accuracy. 

The experimental results are presented in Table~\ref{tab:ablation_ntu60}. The results illustrate the impact of each enhancement strategy on model performance. By progressively incorporating the proposed modules, the model achieves continuous improvement and eventually reaches its optimal results. When using only standard augmentation, the model achieves 71.1\% (X-Sub) and 79.5\% (X-View) accuracy, serving as the baseline. After adding the DW-KRM/JDAM module, the X-Sub accuracy improves to 73.2\%, while the X-View accuracy slightly decreases to 77.1\%. Introducing DGA further enhances both metrics, achieving 77.4\% (X-Sub) and 82.8\% (X-View) — the second-best outcome among all configurations. Finally, when the traditional FIFO queue is replaced by the Entropy-driven Dominance Game Replacement Queue (EDGRQ), the model achieves the best performance, reaching 81.1\% on X-Sub and 89.4\% on X-View. These findings clearly demonstrate that combining all proposed modules yields the most significant performance gains, confirming the effectiveness and synergy of DGA, DW-KRM/JDAM, and EDGRQ within the DoGCLR framework.

\subsubsection{Model Parameter Performance Analysis}

To determine the optimal parameter configuration of the model, experiments are conducted on the NTU RGB+D 60 benchmark dataset using single-modal skeleton data (Joint stream). These experiments aim to select appropriate embedding dimensions and negative-sample queue sizes for the proposed DoGCLR framework.

\textbf{Embedding Dimension.} Table~\ref{tab:embedding_ntu60} presents the linear evaluation results under different embedding dimensions (128, 256, 512, and 1,024). As shown in the results, when the embedding size increases exponentially in powers of two — from 128 to 1,024 — the Top-1 accuracy first rises sharply and then gradually plateaus. Specifically, as the embedding size increases from 128 to 256, the Top-1 accuracy improves significantly from 77.1\% to 89.1\%, indicating a clear performance gain. However, when the embedding size continues to increase from 512 to 1,024, the improvement becomes less pronounced, with Top-1 accuracies of 80.1\% and 83.4\%, respectively. This suggests that the most substantial performance enhancement occurs between 128 and 256 dimensions, while further increases yield diminishing returns, reflecting a potential “saturation” effect. In comparison, the Top-5 accuracy shows a more stable upward trend as the embedding size increases. From 128 to 256, Top-5 accuracy improves from 96.3\% to 98.5\%, but subsequent gains become smaller, particularly between 512 and 1,024, where the values are 94.2\% and 97.2\%, respectively.

Overall, enlarging the embedding dimension exerts a relatively mild impact on Top-5 accuracy, especially when the dimension becomes large. These experimental results demonstrate that the embedding dimension has a significant influence on model performance, particularly at smaller scales (128-256), where the improvement is most pronounced. In practice, selecting the embedding size requires balancing computational cost and model performance. An embedding size of 256 appears to offer a suitable compromise, achieving a desirable balance between accuracy and efficiency.

\begin{table}[t]
\centering
\caption{Ablation Results of the Joint Stream under Different Embedding Dimensions on the NTU RGB+D 60 Dataset (Linear Evaluation)}
\label{tab:embedding_ntu60}
\renewcommand{\arraystretch}{1.15}
\begin{tabular}{|c c c c c|}
\hline
Embedding & 128 & \textbf{256} & 512 & 1,024 \\ \hline
Top-1 & 77.1 & \textbf{89.1} & 80.1 & \underline{83.4} \\ 
Top-5 & 96.3 & \textbf{98.5} & 94.2 & \underline{97.2} \\ \hline
\end{tabular}
\end{table}

\textbf{Negative Sample Memory Bank Size.} Increasing the number of samples stored in the negative sample memory bank typically enhances model performance. To verify this assumption, relevant experiments were conducted, and the results are shown in Table~\ref{tab:memory_ntu60}. According to Table~\ref{tab:memory_ntu60} and the corresponding analyses, expanding the capacity of the memory bank generally leads to improved model performance. In the linear evaluation results on the NTU RGB+D 60 dataset, as the size of the negative sample memory bank increases, both Top-1 and Top-5 accuracies rise steadily. When the memory bank size is 256, the model achieves a Top-1 accuracy of 74.5\%; when the size is increased to 32,768, the Top-1 accuracy improves to 89.1\%, and the Top-5 accuracy rises from 95.3\% to 98.5\%. This demonstrates that increasing the number of negative samples in the memory bank effectively enhances the network’s ability for action recognition. As the memory bank capacity grows, the entropy-based negative sample optimization mechanism can better capture the similarity among intra-class samples, thereby enabling the model to recognize the same actions more accurately across different viewpoints. Hence, utilizing a larger negative sample memory bank deepens the model’s understanding of sample relationships and ultimately leads to a significant improvement in overall performance.

\begin{table}[t]
\centering
\caption{Ablation Results of the Joint Stream under Different Negative Sample Memory Bank Sizes on the NTU RGB+D 60 Dataset (Linear Evaluation)}
\label{tab:memory_ntu60}
\renewcommand{\arraystretch}{1.15}
\begin{tabular}{|c c c c c c c c|}
\hline
Size & 512 & 1,024 & 2,048 & 4,096 & 8,192 & 16,384 & \textbf{32,768} \\ \hline
Top-1 & 77.2 & 75.5 & 78.2 & 74.6 & 78.8 & \underline{84.1} & \textbf{89.1} \\ 
Top-5 & 95.5 & 96.4 & 90.1 & 94.3 & 94.3 & \underline{97.2} & \textbf{98.5} \\ \hline
\end{tabular}
\end{table}


\section{Conclusion}

To overcome the limitations of contrastive learning methods that uniformly process all regions of skeleton sequences, causing motion semantic loss, weak augmentation, and ineffective negative sample selection, this paper proposes a Dominance-Game Contrastive Learning network for skeleton-based action Recognition (DoGCLR). Specifically, a Spatio-temporal Dual-Weight localization algorithm accurately identifies key motion regions by integrating temporal Discrepancy-Degree and spatial Joint-Degree weights. Then, a Dual-scale Dominance Game guides the use of adaptive instance normalization–based style augmentation for key regions and standard augmentation for non-key regions, thereby enriching motion diversity while preserving semantic consistency. Furthermore, an Entropy-driven Dominance Game Replacement Mechanism manages negative samples by evaluating memory banks through information entropy and retaining the most informative ones instead of the conventional FIFO queue. Experiments and ablation studies demonstrate that DoGCLR effectively enhances positive sample quality, increases hard negative diversity, improves key motion modeling, and achieves competitive performance across multiple benchmarks.


%





\ifCLASSOPTIONcaptionsoff
  \newpage
\fi





\bibliographystyle{IEEEtran}
\bibliography{IEEEabrv,Bibliography}
%


\section{Biography Section}
\begin{IEEEbiography}[{\includegraphics[width=1in,height=1.25in,clip,keepaspectratio]{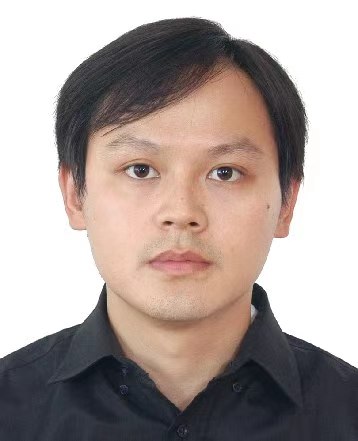}}]{Yanshan Li} received the Ph.D. degree in the South China University of Technology. He is currently a Researcher and Doctoral Supervisor with the Institute of Intelligent Information Processing and Guangdong Key Laboratory of
Intelligent Information Processing, Shenzhen University. His research interests include computer vision, machine learning, and image analysis.
\end{IEEEbiography}

\begin{IEEEbiography}[{\includegraphics[width=1in,height=1.25in,clip,keepaspectratio]{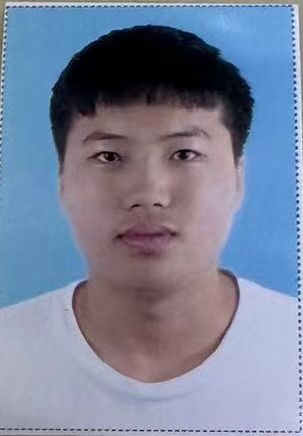}}]{Ke Ma} is currently working toward the master’s student with Shenzhen University. His research focus is on computer vision.
\end{IEEEbiography}

\begin{IEEEbiography}[{\includegraphics[width=1in,height=1.25in,clip,keepaspectratio]{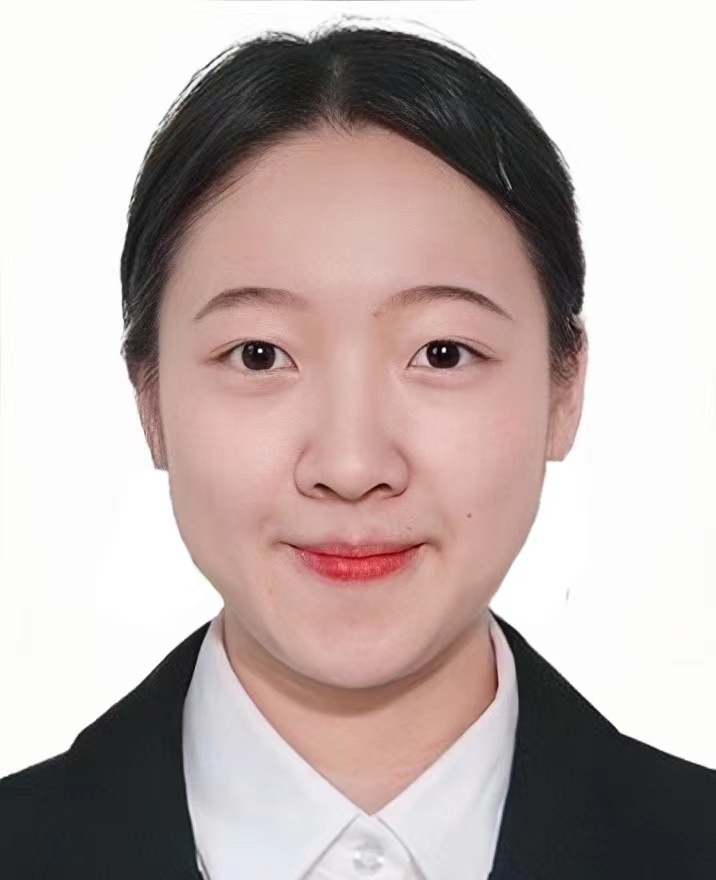}}]{Miaomiao Wei} received her master’s degree from Shenzhen University. Her research mainly focuses on human action recognition.
\end{IEEEbiography}

\begin{IEEEbiography}[{\includegraphics[width=1in,height=1.25in,clip,keepaspectratio]{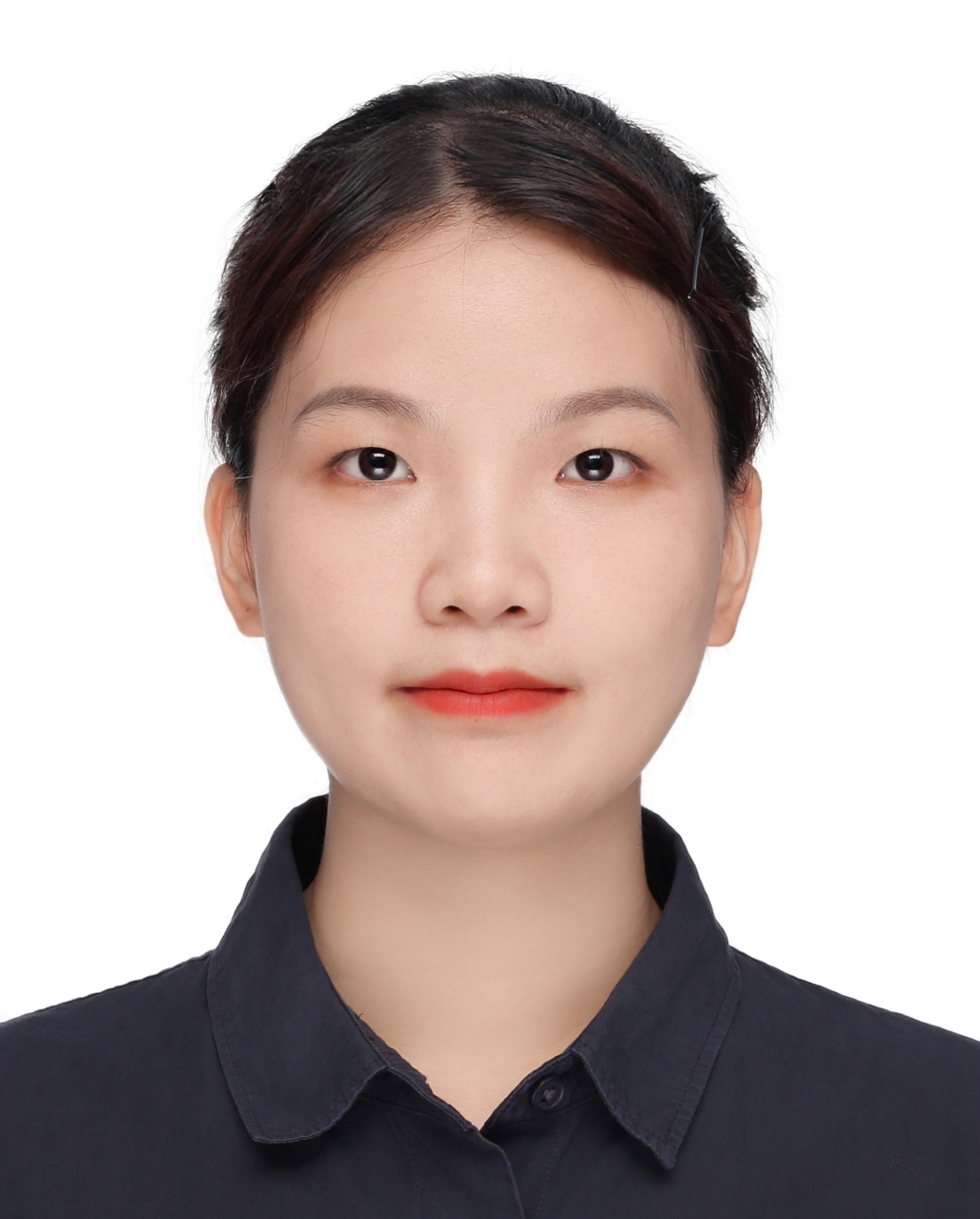}}]{Linhui Dai} (corresponding author) received the Ph.D. degree from the School of Computer Science, Peking University (PKU), China, in 2024, under the supervision of Prof. Hong Liu. She currently serves as an Assistant Professor at the college of Electronics and Information Engineering, Shenzhen University, Shenzhen, China. She has authored or coauthored in PR, T-CSVT, CAAI TRIT, and et al. Her research interests include open world object detection, underwater object detection, and salient object detection. (Email: dailinhui@szu.edu.cn)
\end{IEEEbiography}





\vfill


\end{document}